\documentclass{esannV2}
\usepackage[dvips]{graphicx}
\usepackage[latin1]{inputenc}
\usepackage{epstopdf} 
\usepackage{amssymb,amsmath,array}
\usepackage{url}
\graphicspath{ {./} }
\newlength{\bibitemsep}
\newlength{\bibparskip}
\let\oldthebibliography\thebibliography
\renewcommand\thebibliography[1]{%
  \oldthebibliography{#1}%
  \setlength{\parskip}{\bibitemsep}%
  \setlength{\itemsep}{\bibparskip}%
}
\usepackage{placeins}
\usepackage{float}
\usepackage{eso-pic}

\AtBeginDocument{\AddToShipoutPictureFG*{\AtTextUpperLeft{\put(0,\LenToUnit{40pt}){\parbox{\textwidth}{\centering\bfseries
					27th European Symposium on Artificial Neural Networks, Computational Intelligence and Machine Learning (ESANN), Bruges, Belgium, 2019
}}}}}

\begin{document}
\title{Frequency Domain Transformer Networks for Video Prediction}

\author{Hafez Farazi and Sven Behnke
%
%
\vspace{.3cm}\\
%
University of Bonn, Computer Science Institute VI, Autonomous Intelligent Systems\\
Endenicher Allee 19a, 53115 Bonn, Germany \\
\{farazi, behnke\}@ais.uni-bonn.de
}

\maketitle

\begin{abstract}
The task of video prediction is forecasting the next frames given some previous frames. Despite much recent progress, this task is still challenging mainly due to high nonlinearity in the spatial domain. To address this issue, we propose a novel architecture, Frequency Domain Transformer Network (FDTN), which is an end-to-end learnable model that estimates and uses the transformations of the signal in the frequency domain.  Experimental evaluations show that this approach can outperform some widely used video prediction methods like Video Ladder Network (VLN) and Predictive Gated Pyramids (PGP).
\end{abstract}

\section{Introduction}
In video prediction, the predictor has to model both scene contents and motions. In recent years, deep learning approaches became the first choice for this task.
Although having a deep network which can learn all the aspects of the task by itself is appealing, the history of deep learning shows that an appropriate network structure is key for learning from limited data.
For example, typical properties of images are reflected in the structure of hierarchical convolutional networks.
Video prediction is challenging, due to highly non-linear effects of local translations in the spatial domain.  
Estimating motion and using the estimated motion for prediction is much easier in the frequency domain.
Multiple previous works tried to learn image relations by separating content and transformation~\cite{michalski2014modeling}\cite{memisevic2013learning}. The learned features for these architectures are Gabor-like filters which decompose the signal according to spatial frequency and phase. In the Relational Auto-Encoder (RAE)~\cite{memisevic2013learning}, for example, the paired responses are then multiplied element-wise to estimate transformations between two consecutive frames. We argue that instead of element-wise multiplication of linear filter responses, we can compute the transformation by calculating phase difference in the frequency domain. The estimated phase difference can then easily be used for prediction in frequency space and the predicted frequency representation can be linearly transformed back into the spatial domain. We show the effectiveness of our proposed Frequency Domain Transformer Network (FDTN) approach on three synthetic datasets.

The code and datasets of this paper are publicly available.\footnote{ \url{https://github.com/AIS-Bonn/FreqNet}.}
\section{Related Work}

Although many approaches to the video prediction task have been explored, the most successful approaches utilize deep learning models. Cricri et al.~\cite{cricri2016video} proposed to add recurrent lateral connections in Ladder Networks to capture  temporal dynamics of  video. These recurrent connections, as well as lateral shortcuts, relive the deeper layers from modeling spatial detail. The VLN architecture achieves competitive results to Video Pixel Networks~\cite{kalchbrenner2016video}, the state-of-the-art on Moving MNIST dataset, using far fewer parameters.

Another well-known model is PGP~\cite{michalski2014modeling}, which is based on a gated auto-encoder and the bilinear transformation model of RAE~\cite{memisevic2013learning}. PGP has the assumption that two temporally consecutive frames can be described as a linear transformation of each other. In the PGP model, by using a bi-linear model, the hidden layer of mapping units encodes the transformation. These transformation encodings are then used to predict the next frame. Conv-PGP~\cite{conv-pgp}  reduces the number of parameters significantly, by utilizing convolutional layers.

%

Image registration is a fundamental task in image processing which estimates the relative transformation between two similar images.  A well-known method for image registration using Fourier domain representation is Phase Correlation.  Phase Correlation can be used to calculate the relative translative offset between two similar images. Reddy et al.~\cite{reddy1996fft} demonstrated that rotation and scaling differences between two images can be estimated by converting them to log-polar coordinates. Foroosh et al.~\cite{foroosh2002extension} extended this method to work with subpixel transformation. Sarvaiya et al.~\cite{sarvaiya2012image} proposed an extended version of phase correlation which is more robust and can work under a higher scale. We are inspired by the phase correlation method and designed FDTN.
\vspace{-5px}
\section{Frequency Domain Transformer Networks (FDTN)}

\begin{figure}[t]
\vspace{-5px}
\centering
{\includegraphics[width=0.99\textwidth]{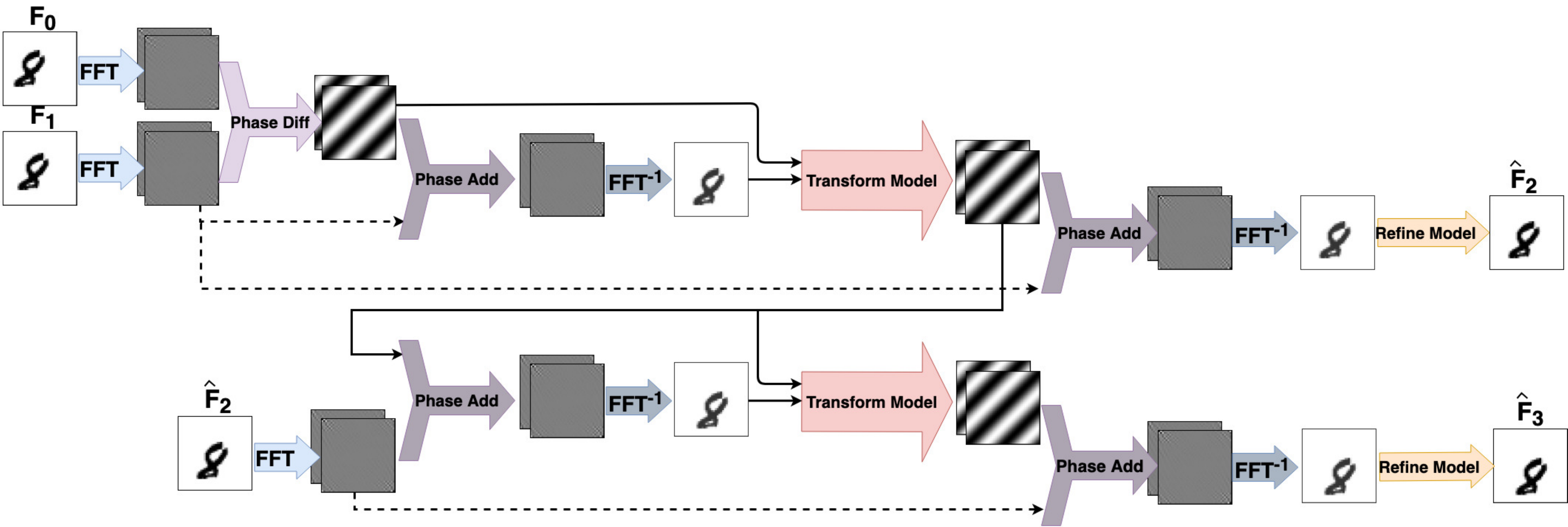}}   
\vspace{-12px}
\caption{The proposed architecture for predicting two frames, given two seed frames. ``Transform Model'' is modifying the encoded transformation for the next prediction. ``Refine Model'' reconstructs details in the spatial domain.}
\label{Teaser}
\end{figure}

If we assume periodic boundary conditions, it is possible to formulate the translation between two consecutive frames as  element-wise differences of the phases of their complex frequency domain representation. We can then use this transformation to predict the next frame in the frequency domain by simple phase addition. The last step is converting the predicted frame to the spatial domain. Fig.~\ref{Teaser} gives an overview of our proposed architecture. 
By using a Transform network in the frequency domain, we relax the periodic boundary assumption.

At the first step, we calculate the Fast Fourier Transform of two seed frames. To obtain the translation between two consecutive frames, we calculate the element-wise phase difference of those frames in frequency domain:
\vspace{-5px}
\begin{equation} \label{eqHFREL}
\vspace{-7px}
\begin{split}
R^{i,j}=\begin{bmatrix}\Re{(R^{i,j})}\\\\ \Im{(R^{i,j})} \end{bmatrix}=\begin{bmatrix}cos(H_0^{i,j},H_1^{i,j})\\\\sin(H_0^{i,j},H_1^{i,j})\end{bmatrix}=\begin{bmatrix}\frac{H_0^{i,j}.H_1^{i,j}}{||H_0^{i,j}||.||H_1^{i,j}||+\varepsilon}\\\\\frac{H_0^{i,j}\times{H_1^{i,j}}}{||H_0^{i,j}||.||H_1^{i,j}||+\varepsilon}\end{bmatrix},
\end{split}
\end{equation}
where $H_k^{i,j}$ is a vector in the complex plane of the Fourier domain for the $Frame_k$. $R^{i,j}$ is encoding the transformation in the Fourier domain which has the shape of $[W\times{H}\times{2}]$, while each frame has the shape of $[W\times{H}]$. Note that a small positive constant $\varepsilon$ is added for numerical stability.

It is possible to encode higher-order transformations like acceleration by calculating the difference of differences using Eq.~\ref{eqHFREL}. It is also possible to filter the noise in the $R^{i,j}$ by utilizing multiple observations.

We passed the transformation representation $R^{i,j}$ to ``Transform Model'', a feed-forward network, to address the changes of the transformation. This model will change the $R^{i,j}$ in a way that it is suitable for the next frame prediction.
Then, we use the transformed $R^{i,j}$ for predicting the next frame in the Fourier domain. We rotate each element by using the constructed rotation matrix:
\vspace{-5px}
\begin{equation} \label{eqHFP}
\vspace{-3px}
\begin{split}
\hat{H_t^{i,j}}=\begin{bmatrix}\Re{(H_{t}^{i,j})}\\\\ \Im{(H_{t}^{i,j})} \end{bmatrix}=\begin{bmatrix}\Re{(R^{i,j})}&-\Im{(R^{i,j})}\\\\\Im{(R^{i,j})}&\Re{(R^{i,j})}\end{bmatrix}\begin{bmatrix}\Re{(H_{t-1}^{i,j})}\\\\ 
\Im{(H_{t-1}^{i,j})} \end{bmatrix},
\end{split}
\end{equation}
where $\hat{H_t^{i,j}}$ is the prediction of the next frame in Fourier domain. We can then obtain the predicted frame in time domain using the inverse FFT.

Although after inverse FFT we have the predicted frame, due to some numerical imprecisions, the result can become blurry after long prediction. This can be mitigated using the ``Refine Model'', another feed-forward network that is designed for reconstructing detail in the spatial domain.
\section{Experimental Results}
\subsection{Datasets}

We used three different datasets to evaluate our proposed architecture.
\textit{Moving Morse Code} is a simple one-dimensional dataset that contains patterns, which are chosen randomly from 36 different Morse codes. The patterns are moving with a random constant velocity. \textit{Moving MNIST} contains ten frames with one MNIST digit moving inside a 40$\times$40 frame. Digits are chosen randomly from training and test set and placed at a random position with a random velocity. \textit{Bouncing Ball} dataset contains ten frames with one round object moving inside a 40$\times$40 frame. Balls are positioned randomly with a random velocity. Note that the ball can move with subpixel velocity.

\subsection{Models}
We used two different trainable models in our computational graph. Each has a different purpose. ``Transform Model''  is designed to change the transformations between frames. In Moving MNIST and Bouncing Ball, this model is responsible for changing the motion of digits.
We propose two Transform Model versions: FDTN(FC) and FDTN(Conv). FDTN(FC) is utilizing two fully-connected layers with sigmoid activations. FDTN(Conv) is designed to utilize the structure of the data by using a three-layer convolutional network with ReLU activations. The convolutional version is more efficient than the fully connected version, and it has fewer parameters. The only issue using convolutional layers is the fact that due to the location-invariant nature of convolutions, we cannot model location-dependent features. To address this issue, we used location-dependent convolutional layers, proposed by Azizi et al.~\cite{AziziFarazi2018}. Fig.~\ref{longRes}(f) shows that if we eliminate the Transform Model we cannot predict the changes of transformations.

To mirror the velocity at the border in 2-dimensional datasets, we can flip $R^{i,j}$  around the desired axes. We calculate four different versions of $R^{i,j}$; the original $R^{i,j}$, flipped vertically, horizontally, and both. In the last part of the ``Transform Model'' network, we have a softmax layer, which can weight between these four different versions. The weighted sum is then routed for Phase Adding operation to calculate the next frame. The input to the model is the predicted frame without the ``Transform Model'' applied. Note that to have a more efficient inference implementation, if the object does not need to change transformation, we can route the predicted frame directly to the ``Refine Model''.
The implementation of the ``Transform Model'' for the Morse Code dataset is different. Since we don't need to change the velocity at the border, we denoise $R^{i,j}$ using fully connected layers.

\begin{figure}[b]
\centering
{\includegraphics[clip,trim=0 470 0 90,width=0.8\textwidth]{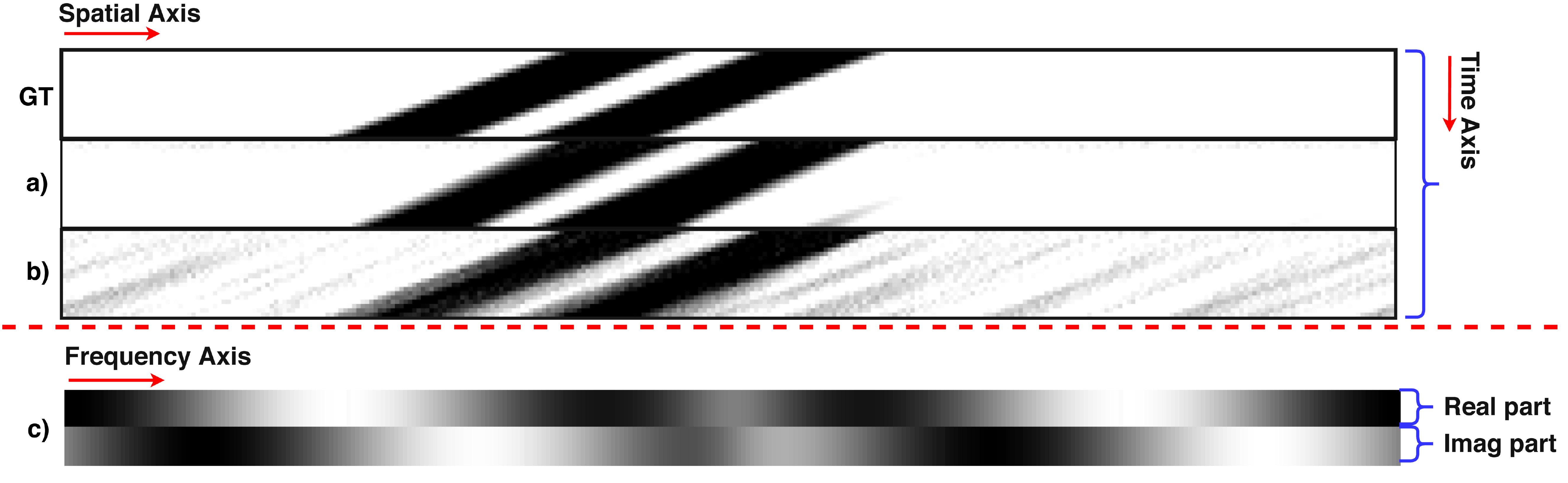}}   
\caption{Moving Morse Code dataset. First two rows are the noisy seeding frames and the rest are predicted using a) FDTN(FC), b) FDTN(FC) without Transform Model; c) Frequency domain representations of transformation $R$.}
\label{FigMMorseCode}
\end{figure}

Due to numerical imprecision, the predicted  frame can become blurry when predicted for a long time. To mitigate this issue, we propose the second learnable model, ``Refine Model'', consisting of three convolutional layers followed by ReLU activations. The result of these is multiplied element-wise to create the output. 
The effect of eliminating this model is shown in Fig.~\ref{longRes}(e).
\subsection{Evaluation}
In our first experiments, we used Moving Morse Code dataset to sanity-check our implementation. A sample result from this dataset is depicted in the Fig.~\ref{FigMMorseCode}.

In Moving Morse Code, we predicted 18 frames from two noisy seed frames. Transform Model can learn to denoise the frequency-domain representation.

\begin{figure}[t]
\centering
{\includegraphics[clip,trim=0 0 0 30,width=0.8\textwidth]{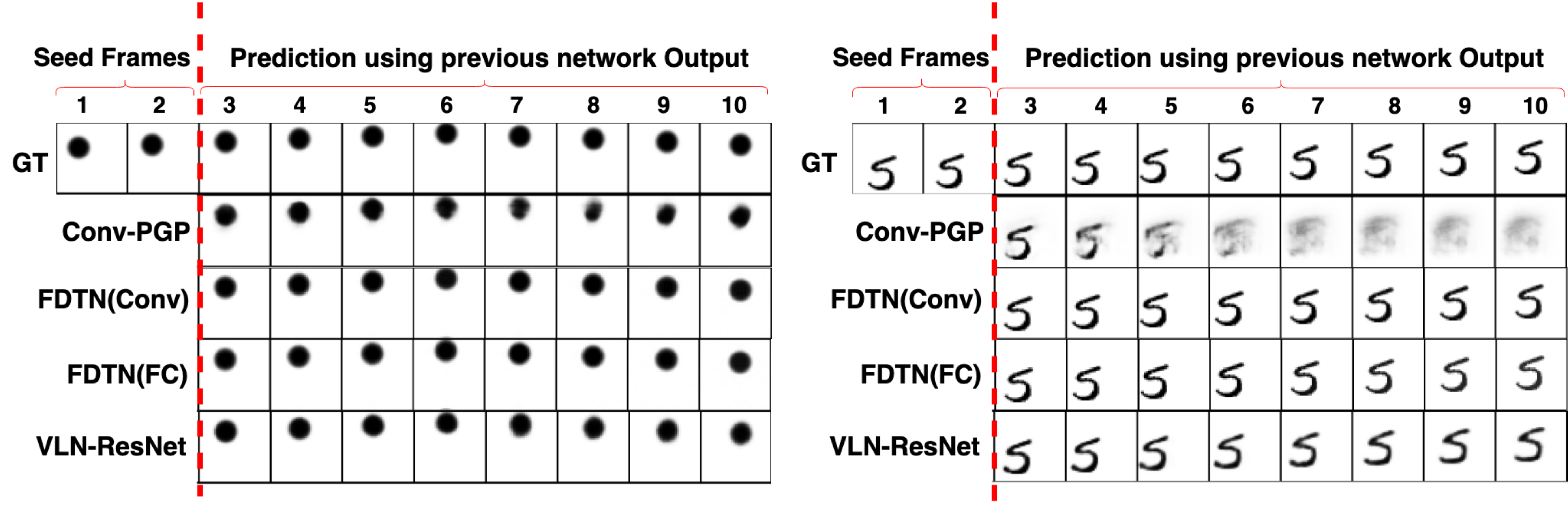}}   
\caption{Predictions for Bouncing Ball and Moving MNIST (random samples).}
\label{SampleRes}
\end{figure}

We evaluate our architecture on both Moving MNIST and Bouncing Ball datasets. We used Conv-PGP and VLN model as the baselines for comparison. In these experiments, we predicted eight frames from two seed inputs. Sample results of our models, as well as used baselines, are presented in Fig.~\ref{SampleRes}. 

Table.~\ref{MNIST_Table_results} reports the prediction loss and the number of parameters for the evaluated models. It can be observed that both of our proposed models outperform our baselines on both Moving MNIST and Bouncing Ball datasets.

\begin{figure}[b]
{\includegraphics[width=0.98\textwidth]{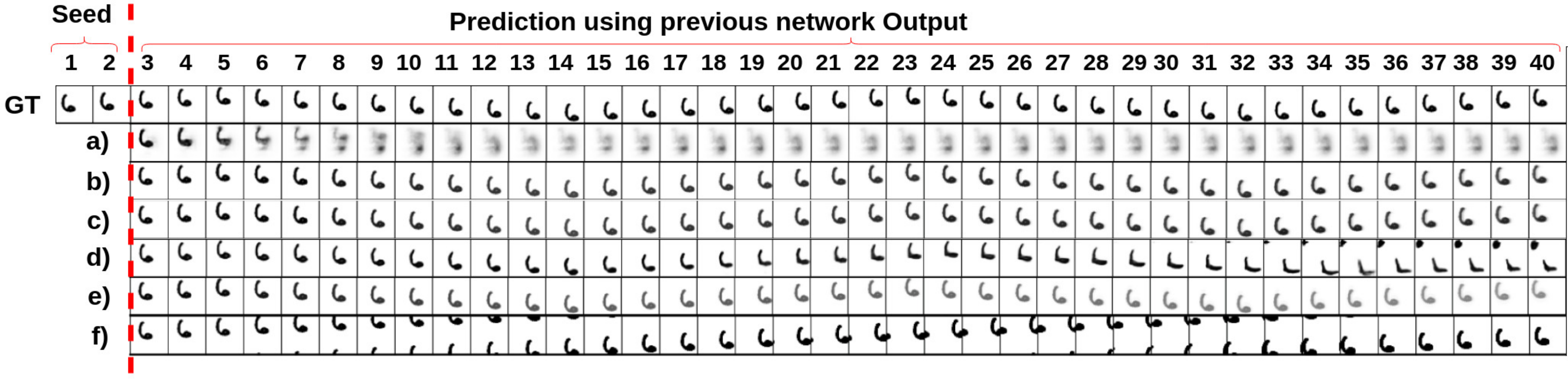}}   
\caption{Moving MNIST models trained for ten predictions and tested on a longer sequence: a) Conv-PGP, b) FDTN(Conv), c) FDTN(FC), d) VLN-ResNet, e) FDTN(FC) without Refine Model, f) FDTN(FC) without Transform Model. Note the effects of Refine Model and Transform Model.}
\label{longRes}
\end{figure}

The model is trained end-to-end using backpropagation through time. We used Adam optimizer and MSE loss. Similar to VLN and Conv-PGP models, at each time-step our method predicts one frame, but in contrast to them our model which is trained for predicting ten sequences, can work well on longer sequences. One sample of longer prediction is presented in Fig.~\ref{longRes}.

\begin{table}[t]
\renewcommand{\arraystretch}{1.1}
\centering
\caption{Mean squared prediction losses for two data sets.}
\vspace{2px}
\begin{tabular}
{l | c | c | c}
 Model\hspace{0px}& \hspace{0px}Moving MNIST\hspace{0px} & \hspace{0px}Bouncing Ball\hspace{0px} &\hspace{0px}Number of parameters\hspace{0px} \\
\hline \hline
Conv-PGP & 0.06963   & 0.00409& 32K\\ 
FDTN(Conv)  & 0.00316   & 0.00092&\textbf{22K}\\
FDTN(FC) & \textbf{0.00285}  & \textbf{0.00086}&160K\\
VLN-ResNet \hspace{3px}& 0.00544 & 0.00107&1.3M\\
\end{tabular}
\vspace{5px}\\
\label{MNIST_Table_results}
\end{table}

\section{Conclusion}
We propose an end-to-end learnable neural network which has a special structure to estimate the transformation between consecutive video frames in frequency domain and use this estimate to make predictions about future frames. Experiments indicate that our proposed architecture can solve video prediction task in synthetic datasets. Our proposed architecture significantly outperforms the results of both VLN and Conv-PGP models on Moving MNIST and Bouncing Ball datasets. The fully connected version performs better than the convolutional one, though with more parameters.

\vspace{-6px}
{\footnotesize \paragraph{Acknowledgment}
\label{acknowledgment}
This work was funded by grant BE 2556/16-1 (Research Unit FOR 2535 Anticipating Human Behavior) of the German Research Foundation (DFG).
}
\bibliographystyle{unsrt}
{\small \bibliography{document}}

\end{document}